\documentclass[letterpaper, 10 pt, conference]{ieeeconf}  
\IEEEoverridecommandlockouts 
\usepackage{amsmath,amsfonts}

\let\labelindent\relax
\usepackage{amssymb, amsthm}

\usepackage{mathtools}

\DeclareMathOperator{\dist}{d}

\usepackage{mydefs,mymath,kernMach,diffge}

\usepackage{calc}
\usepackage{cases}
\usepackage{url}

\usepackage{float,color,graphicx}
\usepackage{tikz}
\usepackage{verbatim}
\usetikzlibrary{calc}
\usetikzlibrary{patterns}
\usepackage{pgfplots}
\pgfplotsset{compat=1.16}

\usepackage{enumitem}
\usepackage[font=footnotesize,skip=4pt]{caption}
\usepackage{cite}
\usepackage{balance}

\usepackage{sgap}

\graphicspath{{nnGeom/}}

\newcommand{\pGap}{\em Potential Gap}
\newcommand{\saferGap}{\em Safer Gap}
\newcommand{\keyhole}{\em Keyhole ZBF}

\title{\LARGE \bf Safer Gap: A Gap-based Local Planner for Safe Navigation with Nonholonomic Mobile Robots}

\author{{Shiyu Feng$^{1,\dagger}$, Ahmad Abuaish$^{2,\dagger}$, Patricio A. Vela$^{2}$}
\thanks{*This work was supported in part by NSF Award \#1849333, by DARPA PAI, and by KACST Fellowship.}%
\thanks{$\dagger$ Equal contribution}
\thanks{$^{1}$S. Feng is with the School of Mechanical Engineering and the School of Electrical and Computer Engineering, Georgia Institute of Technology, Atlanta, GA 30308, USA.
{\tt\small shiyufeng@gatech.edu}}
\thanks{$^{2}$ A. Abuaish , and P.A. Vela are with the School of Electrical and Computer
Engineering and the Institute for Robotics and Intelligent Machines, Georgia Institute of Technology, Atlanta, GA 30308, USA.
{\tt\small \{aabuaish, pvela\}@gatech.edu}}%
}

\begin{document}

\maketitle
\thispagestyle{empty}
\pagestyle{empty}

\begin{abstract}
This paper extends the gap-based navigation technique in {\pGap} by guaranteeing safety for nonholonomic robots for all tiers of the local planner hierarchy, so called {\saferGap}. The first tier generates a B\'{e}zier-based collision-free path through gaps. 
A subset of navigable free-space from the robot through a gap, called the keyhole, is defined to be the 
union of the largest collision-free disc centered on the robot and a trapezoidal region directed through the gap. It is encoded by a shallow neural network zeroing barrier function (ZBF). 
Nonlinear model predictive control (NMPC), with {\keyhole} constraints and output tracking of the B\'{e}zier path, synthesizes a safe kinematically-feasible trajectory. Low-level use of the 
 {\keyhole}  within a point-wise optimization-based safe control synthesis module
 serves as a final safety layer.
 Simulation and experimental validation of {\saferGap} confirm its collision-free navigation properties.
\end{abstract}

\section{Introduction \label{sec:intro}}

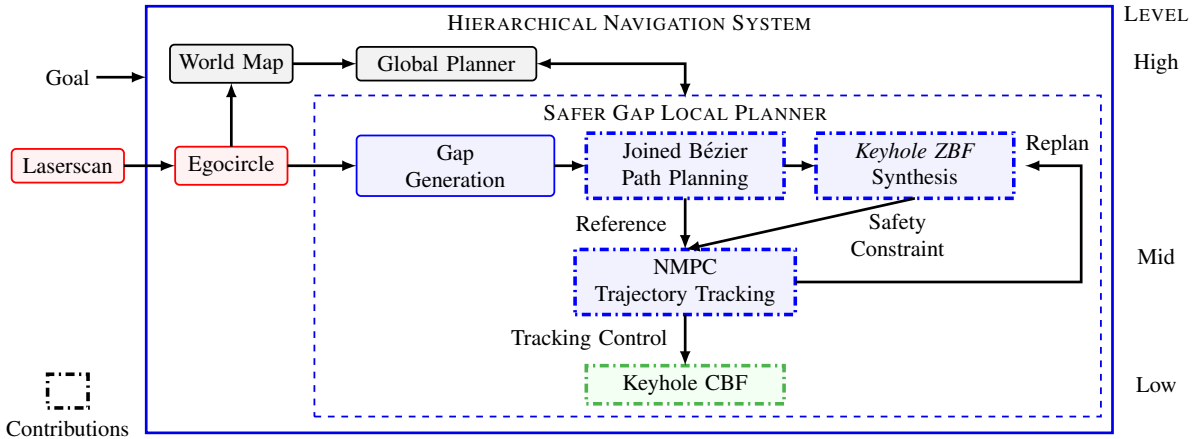
\begin{figure*}[t]
  \vspace*{0.1in}
  \centering
  \hspace*{-0.62in}
  \begin{tikzpicture}[inner sep=0pt,outer sep=0pt]
    \node[anchor=south west] at ($(0, 0)$)
    	{{\scalebox{0.85}{\tikzstyle{block} = [draw, rectangle, text centered, thick,rounded corners=2pt,
                     minimum height=1.5em, minimum width=5em, inner sep=4pt]
\tikzstyle{typical} = [fill=white!95!black]
\tikzstyle{reddish} = [draw=red,fill=white!95!red]
\tikzstyle{blueish} = [draw=blue,fill=white!95!blue]
\tikzstyle{greenish} = [draw=green!40!gray,fill=white!95!green]
\tikzstyle{longblock} = [draw,rectangle,text centered,thick,rounded corners=2pt,
                     minimum height=1.5em, minimum width=8em, inner sep=4pt]
\tikzstyle{largeBlock} = [draw, rectangle, very thick,
                     minimum height=19em, minimum width=43em, inner sep=4pt]
\tikzstyle{smallBlock} = [draw, rectangle, text centered, thick, dashed,
                     minimum height=14.3em, minimum width=35em, inner sep=4pt]
\tikzstyle{dashedBlock} = [draw, dashed, rectangle,
                     minimum height=2em, minimum width=4em, inner sep=4pt]
\tikzstyle{dottedBlock} = [draw, rectangle, text centered, ultra thick,
					 minimum height=1.5em, 
					 minimum width=8em, inner sep=4pt,
					 dash pattern=on 1pt off 2pt on 4pt off 2pt] 
\tikzstyle{newtip} = [->, very thick]
\tikzstyle{bidir} = [<->, very thick]
\tikzstyle{newtip_dashed} = [->, very thick, dashed]
\begin{tikzpicture}[auto, inner sep=0pt, outer sep=0pt, >=latex]

  \node[block, reddish] (laserscan) {Laserscan};

  \node[anchor=south] (goal) at ($(laserscan.north) + (0, 1.0)$) {Goal};
    
  \node[block,reddish,anchor=west] (egocircle) 
    at ($(laserscan.east) + (0.80, 0)$) 
    {\centering Egocircle};
  \node[block, typical, anchor=south] (worldmap)  
    at ($(egocircle.north)+(0, 1)$)
    {\centering World Map};

  \node[longblock, typical, anchor=west] (global) 
    at ($(worldmap.east) + (1., 0)$)  
    {\centering Global Planner};

  \node[longblock, blueish, anchor=west, text width=8em] 
    (gap) at (egocircle-|global.west)
    {\centering Gap \\ Generation};

  \node[dottedBlock,blueish,anchor=west, text width=8em] (traj) 
    at ($(gap.east) + (0.5, 0)$) 
    {\centering Joined B\'{e}zier \\ Path Planning};

  \node[dottedBlock,blueish,anchor=west, text width=8em] (keyhole) 
    at ($(traj.east) + (0.5, 0)$) 
    {\centering {\keyhole} \\ Synthesis};
  
  \node[dottedBlock,blueish,anchor=north, text width=9em] (mpc) 
    at ($(traj.south) - (0, 0.8)$) {\centering NMPC \\ Trajectory Tracking};
    
  \node[dottedBlock,greenish,anchor=north, text width=8em] (controller) 
    at ($(mpc.south) - (0, 0.8)$) {\centering Keyhole CBF};
    
  \node[smallBlock, draw=blue!90!black,anchor=north] (local) 
    at ($(traj)!0.09!(traj) + (1em, 1.1)$){};
  \node[anchor=north,xshift=-1em,yshift=-4pt] (localtext) at (local.north) 
      {\sc Safer Gap Local Planner};

  \node[largeBlock, draw=blue!90!black,anchor=north] (planning) 
    at ($(global)!0.09!(worldmap) + (9em, 0.9)$){};
  \node[anchor=north,xshift=0em,yshift=-4pt] (planningtext) at (planning.north) 
      {\sc Hierarchical Navigation System};

  \draw[newtip] (laserscan.east) -- (egocircle.west);
  \draw[newtip] (egocircle.north) -- (worldmap.south);

  \draw[newtip] (worldmap.east) -- (global.west);
  \draw[newtip] (egocircle.east) -- (gap.west);

  \draw[bidir] (global.east) -- (global.east-|traj.north) -- ($(traj.north)+(0,0.55)$);

  \draw[newtip] (gap.east) -- (traj.west);
  \draw[newtip] (traj.south) -- node[midway,left,anchor=east,xshift=0pt,yshift=0pt,text width=2cm,text centered]{Reference}(mpc.north);

  \draw[newtip] (traj.east) -- (keyhole.west);
  \draw[newtip] (keyhole.south) -- node[midway,right,anchor=west,xshift=15pt,yshift=-5pt,text width=2cm,text centered]{Safety Constraint}(mpc.north);

  \draw[newtip] (mpc.south) -- node[midway,left,anchor=east,xshift=0pt,yshift=0pt,text width=3cm,text centered]{Tracking Control}(controller.north);
  
  \draw[newtip,<-] ($(keyhole.east)+(5pt,0)$) -- node[midway,above,anchor=south,xshift=2pt,yshift=5pt]{Replan}($(keyhole.east)+(30pt,0pt)$) |- (mpc.east);

  \draw[newtip] ($(goal.east)+(3pt,0pt)$) -- (goal.east-|planning.west);
%
%

  \node[anchor=north west, xshift=5pt] (LT) at (planning.north east)
   {\sc Level};
  \node[anchor=center] at (LT|-global) {High};
  \node[anchor=center] at (LT|-local) {Mid};
  \node[anchor=center] at (LT|-controller) {Low};
  
  \node[dottedBlock,anchor=north,minimum width=1em,text width=1em] (legend) at ($(laserscan.south) + (0, -3)$) {};
  \node[anchor=north] (legend_name) at ($(legend.south) + (0, -0.2)$) {Contributions};
\end{tikzpicture}}}};
  \end{tikzpicture}%
  \caption{Hierarchical navigation system with Safer Gap local planner. Red blocks are perception module to generate egocircle. Blue blocks are planning modules. Green block is the control module.\label{fig:pipeline}}
  \vspace*{-1.1em}
\end{figure*}





The local planner module in a hierarchical navigation system processes sensory data to perceive the local environment and represent it as needed by planning and control algorithms. Gaps are such representations, which were shown to support collision-free navigation of idealized robots based on artificial potential fields (APF)
\cite{pgap}. The method, called \textit{Potential Gap} generates safe trajectories guiding the robot through its ego-centric free space toward local gap-based goals. Safety guarantees hold for point-mass holonomic robots
but not for nonholonomic robots. Additional robustification mechanisms were added to improve forward invariance of the  safe space prescribed by the APF. Formal safety guarantees were lost and passage through gaps was compromised.  This paper extends gap-based local planning safety guarantees to nonholonomic models. 

Safety relies on adequately modeling the collision-free space of the robot. Fig.~\ref{fig:bezier_safe} depicts such a region specialized to the gap between two obstacles. It is represented by the union of the largest robot-centered disc and the region between the gap lines. Also depicts is the inflated free area, in dark gray, due to the robot's physical footprint. To use zeroing barrier function (ZBF) safe control synthesis, the free space must be represented by a continuously differentiable implicit function positive inside the safe region and negative outside (the zero level-set defines the boundary). As safe region resembles the shape of a ``keyhole'', we will call the ZBF a {\keyhole}.

The hierarchical navigation system, depicted in Fig.~\ref{fig:pipeline}, relies on a global planner to generate a candidate path to the goal in the world map. The {\saferGap} local planner reacts to local unknown environments and achieves obstacle avoidance. First, a smooth collision-free path based on B\'{e}zier curves is generated from detected gaps. 
If the B\'{e}zier curve control points are contained in the {\keyhole} boundary, then the curve is also contained in the same region \cite{bezierbook}; thus, the path is guaranteed to be collision-free. 
Nonlinear model predictive control (NMPC) is applied to track that path under kinematic feasibility constraints. To guarantee safe path tracking, the {\keyhole} is transcribed as a hard constraint in NMPC. Finally, as a last measure of safety to cope with the rapidly changing environment and the slower rate of NMPC, the {\keyhole} is enforced in a control BF (CBF) point-wise optimization safe control synthesis.

The contributions and organization of {\saferGap} as annotated in Fig.~\ref{fig:pipeline} (dashed) are: \textbf{(i)} Joined B\'{e}zier path generation (\S \ref{sec:bp}) guarantees safety and kinematic passibility through gaps. \textbf{(ii)} The real-time synthesized {\keyhole} (\S \ref{sec:keyhole}) models the safe keyhole region.  NMPC trajectory tracking (\S \ref{sec:nmpc}) with {\keyhole} constraint and nonholonomic dynamics synthesizes safe trajectories. \textbf{(iii)} The {\keyhole} acts as a CBF (\S \ref{sec:cbf}) to ensure safety at the lowest level. 
Simulation benchmarking and real experiments in \S \ref{sec:exp} confirm {\saferGap}'s collision-free properties.

\section{Related Work \label{sec:back}}
\subsection{Vision-based Planning}



Vision-based navigation in unknown environments has become popular nowadays. Perception acts as the first module in navigation frameworks. It is important to efficiently and accurately describe environments for safe path planning and control. Perception can be generally categorized into allocentric and ego-centric approaches \cite{Smith2020}. Within the human's neural hierarchy, ego-centric processing usually happens before allocentric estimation. It needs fewer memory resources and is computationally efficient in depicting the local region. 

Gap is one of the ego-centric methods that can model locally free space and leverage line-of-sight visibility. Although there is no formal definition of the ``gap", it is usually a segment of 1D laserscan measurement, and consists of starting and ending points to represent the collision-free region \cite{followthegap,CG,TCG,SG,AG,pgap,SmEtAl[2020]egoTEB}. Path planning with gaps should be able to improve safety and passibility. Especially, {\pGap} local planner \cite{pgap} proves safety guarantee for point-mass holonomic robot models. It includes radial extension and projection operator to enhance safety for nonholonomic mobile robots. However, the closed-form proof does not hold. 

After modeling environments, the next step is planning a collision-free path within the perception space. Dijkstra’s algorithm, A$^\star$, D$^\star$, D$^\star$ Lite, and AD$^\star$ \cite{dijkstra,Astar,Dstar,DstarLite,LIKHACHEV20081613} are graph search planning methods in costmaps. PRM, RRT, and RRT-X \cite{prm,lazyPRM,rrt,Otte2014RRTXRM} are sample-based planners that search for collision-free paths in the workspace. These are generally used as global planers. Reactive policies such as EB, DWA, TEB and egoTEB \cite{eb,DWA,TEB,SmEtAl[2020]egoTEB} generate local plans to avoid obstacles and approach to goals. 

In addition, artificial potential field (APF) offers fast computation for obstacle avoidance \cite{APF1985,APF_numerical,APF_newfunction, APFSimAnnel, APFRegression}. {\pGap} local planner \cite{pgap} uses APF to synthesize safety guaranteed paths for point-mass holonomic models.
Also, B\'{e}zier curve is a well-known technique to generate smooth trajectories for different robot models \cite{Choi2008PathPB,ELHOSENY2018339,s21072460,Zheng2020BezierTraj,HASSANI2018305,BryceUAVBezier}. 
Our recent work, B\'{e}zier Gap \cite{bgap}, leverages the property of B\'{e}zier curve to synthesize safe trajectories for holonomic quadrupedal robots. However, there is no safety guarantee for nonholonomic models in planning. Consequently, it motivates the research to leverage the advantages of gaps and B\'{e}zier curves for nonholonomic path planning.

\subsection{Nonlinear Model Predictive Control}

Model predictive control (MPC) is a receding-horizon controller that can be used for both generating or tracking trajectories\cite{mpc-book}. For the latter, trajectories are typically generated by a different module that uses a simplified model of the system to allow for fast trajectory generation. In particular, \cite{mpc-wheeled,mpcTEB} demonstrates the efficacy of NMPC for path and trajectory tracking for nonholonomic robots, supporting the utility of NMPC for local planning. Further, enforcing safety constraints in a MPC fashion leads to unconservative safe trajectories. 

\subsection{Safety and Control}

In recent years, safety in control systems has been incorporated via invariant set theory analysis. To ensure safety for a particular set, e.g., free space set $\freeSpace$ in navigation applications, one needs to prove that $\mathcal{F}$ is an invariant set. The safe set is typically represented by the zero sublevel set of a continuously differentiable implicit function, termed barrier function, $h(\tPose):\Real^n\rightarrow\Real$.
\begin{align}
    \mathcal{F} &= \setof{\tPose\in\Real^n}{h(\tPose)\geq 0} \\ 
    \partial\mathcal{F} &= \setof{\tPose\in\Real^n}{h(\tPose)= 0}
\end{align}
where, $\partial\mathcal{F}$ denotes the boundary of the set.

Further, for controlled systems, the control barrier function (CBF) is introduced, for which the control action of the system must render the safe set invariant. The traditional implementation of CBF-based control takes place at the last tier of the control system hierarchy in the form of a point-wise optimization problem that seeks to synthesize safe control actions by satisfying the CBF constraint while minimally deviating from the base controller output \cite{AA17TP}. If the system dynamics are control affine, the aforementioned optimization problem will take the form of a quadratic program (QP).
\begin{equation}  \eqlabel{cbf-qp}
\begin{aligned}
    \min_{\tControl} \quad &\norm{\tControl-\tControl_r}^2\\
    \text{s.t.} \quad & \nabla h^T(\tPose)f(\tPose) + \nabla h^T(\tPose)g(\tPose)\tControl \geq -\gamma h(\tPose)\\
                \quad & \tControl_{min} \leq \tControl_i \leq \tControl_{max}, \, \forall i=1,\cdots,m
\end{aligned}
\end{equation}
where, $\tControl_r\in \Real^m$ is the output of the base controller, $f(\tPose):\Real^n\rightarrow \Real^n$ and $g(\tPose):\Real^n\rightarrow\Real^{n\times m}$ are the system dynamics matrices, $\gamma$ tunable positive parameter, and $\tControl_{min}$ and $\tControl_{max}$ are actuation limits.

However, CBF-based full safe trajectory synthesis has been investigated in \cite{HM22ET} and \cite{JZ21KS}. In the former, the CBF constraint is elevated into a state differential equation facilitating the utilization of traditional control techniques, e.g., LQR and pole placement. In the latter, a discrete form of the CBF constraint in \eqref{cbf-qp} is added to a discrete NMPC formulation. But the resulting safe trajectories are similar to those obtained from discrete NMPC with BF constraint with inflation of the boundaries. As a result, in this work, the {\keyhole} will be added as a discrete position constraint in NMPC formulation for trajectory tracking, as will be discussed in the next section.

To guarantee the existence of solutions to CBF-QP in \eqref{cbf-qp}, $h(\tPose)$ must be a valid CBF, which is usually difficult to certify and can only be done offline. Sum-of-squares is one technique to solve for the polynomial coefficients of $h(\tPose)$ \cite{AAsos}. Similarly, Hamilton Jacobi (HJ) based reachability analysis can be used to generate a safe backward-reachable-set, which can be used as a CBF \cite{SB17CT,MC18CT}. Since both of those methods require offline computation, it is difficult to deploy them in changing environments, although some techniques have been proposed to synthesize safe trajectories from a library of offline-computed reachable sets \cite{Kousik2020,FasTrack}.


\section{{\saferGap} Local Planner \label{sec:safe}}

\begin{figure}[t]
  \vspace*{0.065in}
  \centering
  \scalebox{0.75}{
  \begin{tikzpicture}[inner sep=0pt,outer sep=0pt]
    \node[anchor=south west] at ($(0, 0)$)
    	{{\input{figs/bezier_path}}};
  \end{tikzpicture}%
  }
  \caption{B\'{e}zier trajectory synthesis. $\pose(0)$ is the robot origin. $\eleInd{\cpf}{1}$ and $\eleInd{\cpf}{2}$ are the second and third control points for the first cubic B\'{e}zier curve. Blue lines $\gapSide_l$ and $\gapSide_r$ are left and right gap sides. Red circle is the largest circular free space in egocircle. $\point_l$ and $\point_r$ are the left and right intersection points of $\gapCirc$ and gap sides. Local waypoint $\goal$ is inside the inflated safe region $\infFreeSpace$ to guarantee safety. $\circPt$ is the goal biased point on $\infGapCirc$. Dashed lines show the B\'{e}zier polygons. The combination of brown and cyan paths is the final synthesized path. \label{fig:bezier_safe}}
\end{figure}

This section introduces a gap-based local planning policy to guarantee safe navigation for nonholonomic mobile robots, so called {\saferGap}. It incorporates line-of-sight visibility from gap detection to construct collision free space. Then safety and passibility are maintained during the design of path planning and motion control. 

\subsection{Joined B\'{e}zier Path Planning \label{sec:bp}}

In our previous work \cite{pgap,bgap}, gap-based perception space and B\'{e}zier-based trajectory synthesis are demonstrated to have good navigation performance. However, safety is only guaranteed for the holonomic robot model. We propose to synthesize smooth and safe paths from gaps based on joined B\'{e}zier curves. 
Gap depicts the open region between two obstacles by considering robot's line-of-sight visibility, as shown in Fig.~\ref{fig:bezier_safe}. It is generated from egocircle and follows the same procedure in \cite{pgap}. The egocircle \cite{Smith2020} is an ego-centric 1D array that contains spatial and temporal information of the environment. 

\subsubsection{Collision-free space generation}
Similar to \cite{bgap}, a collision-free space $\freeSpace$ is geometrically constructed for each gap. However, triangle regions constructed in \cite{bgap} are too compact when one side of the robot is close to an obstacle; thus, a richer polygonal space should be created. The largest circular collision-free space $\gapCirc$ within egocircle $\egoCirc$ is found, e.g., red circle in Fig.~\ref{fig:bezier_safe}. Two gap points, $\lgap$ and $\rgap$, are initially connected to the tangent points of $\gapCirc$. The raw gap sides are formulated. To be noticed, the tangent point corresponding to $\lgap$ always has smaller polar angle than $\rgap$ in the robot local frame. If raw gap sides are obstructed by other obstacles, inward rotations are applied about $\lgap$ (clockwise) and $\rgap$ (counter clockwise) until there is no obstruction. The maximum rotation can push the tangent point to the center of $\gapCirc$, which constructs a minimum collision free space $\freeSpace$, which is the same as \cite{bgap}. After rotation, the gap sides, $\gapSide_l$ and $\gapSide_r$ are finalized. We define the intersections between $\gapSide$ and $\gapCirc$ are $\point_l$ and $\point_r$. Two gap points and two intersected points formulate a collision free polygon $\gapPoly$. The full collision free space $\freeSpace$ is 
\begin{equation}
    \freeSpace = \gapPoly \cup \gapCirc
\end{equation}
Considering robot geometry, $\freeSpace$ is inflated as $\infFreeSpace$ in Fig.~\ref{fig:bezier_safe}. The inflation size is a function of robot radius. Intersected points after inflation are denoted as $\inflate{\point_{l}}$ and $\inflate{\point_{r}}$. Inflated gap circle is $\infGapCirc$. Any path $\bPath$ within the inflated collision free space $\bPath \in \infFreeSpace$ can guarantee safety for the full robot geometry.

\subsubsection{Joined B\'{e}zier curves}
The first segment is a cubic B\'{e}zier curve parameterized by $u$ inside $\infGapCirc$,
\begin{equation} \label{eq:cb}
\begin{aligned}
  \bezierCurve_1(u) &= \sum_{i=0}^{n=3} \binom{n}{i} (1-u)^{n-i} u^i \eleInd{\cpf}{i} \\
  \binom{n}{i} &= \frac{n!}{i!(n-i)!}, \; 0 \leq u \leq 1
\end{aligned}
\end{equation}
where $\eleInd{\cpf}{i}$ is the $i$th control point of $\bezierCurve_1$.

Since the gap is detected in robot local frame, robot center is used as the first control point $\eleInd{\cpf}{0} = \pose(0)$. An intermediate point $\circPt$ is defined on the arc between $\inflate{\point_l}$ and $\inflate{\point_r}$, and served as the last control point $\eleInd{\cpf}{3}$. The other two control points are designed from initial orientation $\theta(0)$, linear velocity $\nu(0)$, and acceleration $\acc(0)$ of the nonholonomic robot. $\eleInd{\cpf}{1}-\eleInd{\cpf}{0}$ is co-linear with the unit orientation vector $\orient(0)=[cos(\theta(0)),sin(\theta(0))]$. Curve velocities and accelerations are obtained from the first and second derivatives of cubic B\'{e}zier curve, 
\begin{align}
 \dot{\bezierCurve_1}(u) &= 3 \sum_{i=0}^{2} \binom{2}{i} (1-u)^{2-i} u^i (\eleInd{\cpf}{i+1}-\eleInd{\cpf}{i})  \\
  \dot{\bezierCurve_1}(0)&=3(\eleInd{\cpf}{1}-\eleInd{\cpf}{0}) \\
  \ddot{\bezierCurve_1}(u) &= 6 \sum_{i=0}^{1} \binom{1}{i} (1-u)^{1-i} u^i (\eleInd{\cpf}{i+2}-2\eleInd{\cpf}{i+1}+\eleInd{\cpf}{i})  \\
  \ddot{\bezierCurve_1}(0)&=6(\eleInd{\cpf}{2}-2\eleInd{\cpf}{1}+\eleInd{\cpf}{0}).
\end{align}

The curve parameter $u \in [0,1]$ should be scaled to time $t \in [0,\timeScaleF]$ and $t = \timeScaleF u$. The final time $\timeScaleF$ is estimated by $||\circPt-\pose(0)||/\nu_d$, where $\nu_d$ is the robot desired linear velocity. Then the scaled B\'{e}zier path $\scale{\bezierCurve}_1(t)=\bezierCurve_1(t/\timeScaleF)$, and
\begin{align}
    \dot{\scale{\bezierCurve}_1}(t) &= \frac{1}{\timeScaleF} \dot{\bezierCurve_1}(\frac{t}{\timeScaleF}) \\
    \ddot{\scale{\bezierCurve}_1}(t) &= \frac{1}{\timeScaleF^2} \ddot{\bezierCurve_1}(\frac{t}{\timeScaleF}).
\end{align}

Set $||\dot{\scale{\bezierCurve}_1}(0)||=\nu(0)$, which needs $||\eleInd{\cpf}{1}-\eleInd{\cpf}{0}||=\timeScaleF \nu(0)/3$. Then set $\ddot{\scale{\bezierCurve}_1}(0)=\acc(0)$, all control points for the first B\'{e}zier path segment $\bezierCurve_1(u), u \in [0,1]$ are uniquely defined
\begin{equation}
\begin{aligned}
    \eleInd{\cpf}{0} &= \pose(0) \\
    \eleInd{\cpf}{1} &= \pose(0) + \frac{\timeScaleF \nu(0)}{3}\orient(0) \\
    \eleInd{\cpf}{2} &= \frac{\timeScaleF^2}{6} \acc(0) - \eleInd{\cpf}{0} + 2 \eleInd{\cpf}{1} \\
    \eleInd{\cpf}{3} &= \circPt
\end{aligned}
\end{equation}

The second path segment is generated from a quadratic B\'{e}zier curve
\begin{equation} 
  \bezierCurve_2(u) = (1 - u)^2 \eleInd{\cps}{0} + 2(1 - u)u \eleInd{\cps}{1} + u^2 \eleInd{\cps}{2}.
\end{equation}
where $\eleInd{\cps}{0}=\circPt$. 

G1 continuity maintains a smooth connection between B\'{e}zier curves. Therefore, the direction vector $\directVec$ should satisfy the equality:
\begin{equation} 
  \directVec = \frac{\eleInd{\cps}{1}-\eleInd{\cps}{0}}{||\eleInd{\cps}{1}-\eleInd{\cps}{0}||} = \frac{\eleInd{\cpf}{3}-\eleInd{\cpf}{2}}{||\eleInd{\cpf}{3}-\eleInd{\cpf}{2}||}
\end{equation}
The magnitude of $\eleInd{\cps}{1}-\eleInd{\cps}{0}$ is calculated by desired linear velocity $\nu_d$. With quadratic B\'{e}zier curve and similar scale mechanism,
\begin{align}
  \dot{\bezierCurve_2}(u) &= 2(1-u)(\eleInd{\cps}{1} - \eleInd{\cps}{0}) + 2u(\eleInd{\cps}{2} - \eleInd{\cps}{1}) \\
  \dot{\bezierCurve_2}(0) &= 2(\eleInd{\cps}{1} - \eleInd{\cps}{0}) \\
  \dot{\scale{\bezierCurve}_2}(t) &= \frac{1}{\timeScaleS} \dot{\bezierCurve_2}(\frac{t}{\timeScaleS})
\end{align}
where $\timeScaleS=||\goal-\circPt||/\nu_d$. 

Similarly, set $||\dot{\scale{\bezierCurve}_2}(0)||=\nu_d$, which requires $||\eleInd{\cps}{1}-\eleInd{\cps}{0}||=\timeScaleS \nu_d / 2$. When $\circPt$ is close to $\inflate{\gapSide_l}$ or $\inflate{\gapSide_r}$, $\eleInd{\cps}{1}$ is possible to be outside of the inflated gap sides after scaling. A length scale number $\lambda \in (0,1]$ is calculated to bound $\eleInd{\cps}{1}$ inside $\infFreeSpace$. All control points for the second B\'{e}zier path segment $\bezierCurve_2(u)$ are constrained
\begin{equation}
\begin{aligned}
    \eleInd{\cps}{0} &= \circPt \\
    \eleInd{\cps}{1} &= \circPt + \lambda \frac{\timeScaleS \nu_d}{2}\directVec \\
    \eleInd{\cps}{2} &= \goal \\
\end{aligned}
\end{equation}

Local waypoint $\goal$ candidates are initially found based on global plans and then bounded by $\infFreeSpace$. The intermediate point $\circPt$ starts with the middle point of the arc, then is biased by the relative position between $\pose(0)$ and $\goal$ to synthesize smoother paths. If $\goal$ is within $\infGapCirc$, only first B\'{e}zier segment is computed. The final B\'{e}zier-based path is 
\begin{equation}
    \bPath(u) = \begin{cases}
        \bezierCurve_1(u), & \goal \in \gapCirc \\
        \bezierCurve_1(u) \cup \bezierCurve_2(u), & \text{otherwise}
    \end{cases}
\end{equation}

From the above design, the first B\'{e}zier polygon for $\bezierCurve_1(u)$ is always within $\infGapCirc$. The second B\'{e}zier polygon is within the convex region $\inflate{\gapPoly}$.
Therefore, the joined B\'{e}zier path is inside the inflated collision free space, $\bPath(u) \subseteq \infFreeSpace$. Safety and passibility are achieved for nonholonomic robots.
It only takes $\leq 2ms$ to generate path for each gap (on Intel i7-8700). The full path planning time depends on the number of detected gaps. A set of new paths are synthesized in every planning loop.

\subsubsection{Path scoring}
\begin{figure}[t]
 \vspace*{0.065in}
  \centering
  \scalebox{0.75}{
  \begin{tikzpicture}[inner sep=0pt,outer sep=0pt]
    \node[anchor=south west] at ($(0, 0)$)
    	{{\includegraphics[height=2.8in]{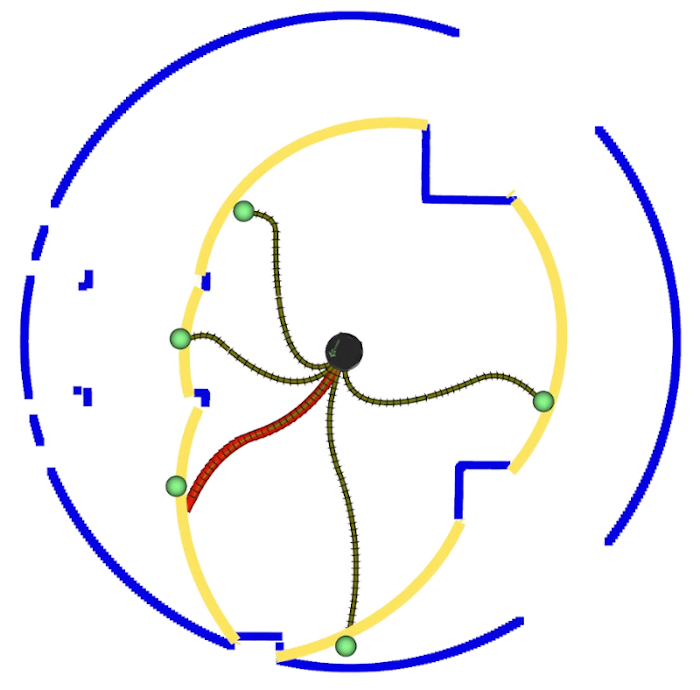}}};
  \end{tikzpicture}%
  }
  \caption{Joined B\'{e}zier paths for all gaps. Blue is egocircle $\egoCirc$. Yellow are 5 detected gaps. Green points are local waypoints $\goal$. Black paths are the synthesized B\'{e}zier paths $\bPath$. Red path is the selected $\bPath^*$ based on the scoring equation. \label{fig:bezier_path}}
  \vspace*{-1.5em}
\end{figure}

A scoring function is computed for each joined B\'{e}zier path to choose the best executed one $\bPath^*$. This function is an improved version from \cite{pgap} by adding an orientation cost. The path has lower deviation from robot's orientation is preferable, since nonholonomic robots cannot suddenly turn backwards. It is also helpful to pick the correct path when the final goal point is on the other side of walls.
\begin{multline*}
    \nonumber 
    J(\bPath) = \sum_{\tPose \in \bPath} 
        C(\dist(\tPose,\egoCirc)) + w_1||\tPose_\text{end} - \pose^*|| 
       + w_2 |\theta_\text{end} - \theta(0)| 
\end{multline*}
\vspace{-1.em}
\begin{eqnarray}
  \nonumber 
  \small{\text{where} \quad
    C(d) = \begin{cases}
      c_{\text{obs}} e^{-w_2 (d - \rIns)}, & d > \rIns \\
      0, & d > r_{\text{max}}\\
      \infty, \text{otherwise}
    \end{cases}}
\end{eqnarray}
$\dist(\tPose,\egoCirc)$ is the distance from path pose $\tPose=[\text{x}_1, \text{x}_2, \theta]^T$ to the nearest point on egocircle $\egoCirc$. $||\tPose_\text{end} - \pose^*||$ measures the distance between the end pose of $\bPath$ and the local goal $\pose^*$ from a global plan. $|\theta_\text{end} - \theta(0)|$ is the angle difference between end pose and initial pose. $\rIns$ and $r_{\text{max}}$ are proportional to the robot radius to control the safe distance. $w_1$, $w_2$ and $c_{\text{obs}}$ are tunable weights. Each time, every best path $\bPath^*_i$ compares with the previous executed path $\bPath^*_{i-1}$ to decide whether switching to the new path.
One example is shown in Fig.~\ref{fig:bezier_path}. The best path (red) is selected from a set of B\'{e}zier path candidates.

\newcommand{\trajRef}{\tPose_{\text{ref}}}
\newcommand{\safeFunc}{\mathcal{S}}

\subsection{NMPC Trajectory Tracking \label{sec:nmpc}}

Safe joined B\'{e}zier path $\bPath^*$ is generated in \S \ref{sec:bp}. In order to safely track the path for nonholonomic model, NMPC is applied. Assume the unicycle nonholonomic model  with state $\tPose=[\text{x}_1, \text{x}_2, \theta]^T$ and control $\tControl=[\nu, \omega]^T$. 
\begin{equation}
\begin{aligned}
    \dot{\text{x}}_1 &= \nu cos(\theta) \\
    \dot{\text{x}}_2 &= \nu sin(\theta) \\
    \dot{\theta} &= \omega
\end{aligned}
\end{equation}
In order to assign time stamps to path $\bPath^*$ based on nonholonomic dynamics, near-identity trajectory $\trajRef(t)$ \cite{1025398} is synthesized given the path and desired linear velocity $\nu_d$. Time stamps and the velocity profile $\tControl_{\text{ref}}$ are assigned to the dynamically feasible trajectory reference. However, it is possible to slightly deviate from the original B\'{e}zier path. NMPC with the safety \emph{Keyhole} ZBF constraint can guarantee safety during tracking. The scheme is formulated with initial state $\tPose(t)$ and control $\tControl(t)$ at current time $t$:
\begin{equation}
\begin{aligned}
\min_{\tControl(t+k)} \quad J(t) &= \sum_{k=0}^{N-1} ||\tPose(t+k) - \trajRef(t+k)||_Q \\ 
& \qquad + ||\tControl(t+k) - \tControl_{\text{ref}}(t+k)||_R \\
\textrm{s.t.} \quad & \tPose(t+k+1) = f(\tPose(t+k), \tControl(t+k)) \\
& \tControl_{lb} \leq \tControl(t+k) \leq \tControl_{ub} \\
& \boldsymbol{a}_{lb} \leq |\tControl(t+k+1)-\tControl(t+k)| \leq \boldsymbol{a}_{ub} \\
& h(\tPose(t+k)) \geq 0
\end{aligned}
\end{equation}
where $\norm{z}_Q=z^TQz$, and $\tControl_{lb}$, $\boldsymbol{a}_{lb}$, $\tControl_{ub}$, and $\boldsymbol{a}_{ub}$ are the lower and upper bounds of velocities and accelerations to maintain smooth motions. $N$ is the number of time step in the prediction horizon. $Q$ and $R$ are the state and control weights. $h(\tPose)$ is the {\keyhole}, which represents the inflated collision-free space $\infFreeSpace$.
\subsection{{\keyhole} Synthesis \label{sec:keyhole}}

The inflated collision-free space $\infFreeSpace$ is captured by the zero level-set of the {\keyhole}. We will use a shallow, two-layer neural network with rectified linear units (ReLU) to model the barrier function. In order to keep the network shallow and minimal, we need to leverage the geometry of keyhole shape, i.e., the straight lines and the circle. The complete expression of the {\keyhole} is 
\begin{equation} \eqlabel{nn}
 \begin{split}
    h(x) =\  & \alpha_1 R_1 + \alpha_2 R_2 + \alpha_3 R_3  + \alpha_4 R_c  + \alpha_5 R_1R_2 \\
         & + \alpha_6 R_cR_1 + \alpha_7 R_cR_2  + \alpha_8 R_cR_3  \\
         & + \alpha_9 R_1R_2R_3 + \alpha_{10} R_1R_4R_5  + \alpha_{11} R_2R_4R_5  \\
         & + \alpha_{12} R_cR_1R_4 + \alpha_{13} R_cR_2R_4  + \alpha_{14} R_cR_1R_2 \\
         & + \alpha_{15} R_cR_1R_2R_3  + b
 \end{split}
\end{equation}
Effectively, all points in the domain are mapped onto the level-sets of the line and circle equations (layer 1) and their polynomial combinations (layer 2). Any point that maps onto a negative level-set is set to zero by the ReLU. As shown in \eqref{nn} by the subscripts of $R$, three additional straight lines are added, line 3, 4, and 5. Fig.~\figref{keyhole-lines}-a shows an illustrative example of the keyhole shape with line 3, which connects points $\inflate{\point_l}$ and $\inflate{\point_r}$. Lines 4 and 5 were added to cope with a special keyhole configuration shown in Fig.~\figref{keyhole-lines}-b.
where, $x=[\text{x}_1,\text{x}_2]^T$, $R_i=ReLU(c_i^Tx+d_i)$, $R_c=ReLU(r^2-(x-x_c)^T(x-x_c))$, $ReLU(z)=max(0,z)$, $c_i$ and $d_i$ are the coefficients for the straight lines, and $x_c$ and $r$ are the center and radius of $\infGapCirc$, respectively.

\begin{figure}[t]
  \vspace*{0.06in}
  \centering
  \scalebox{0.75}{
  \begin{tikzpicture}[inner sep=0pt,outer sep=0pt]
    \node[anchor=south west] at ($(0, 0)$)
    	{{

\begin{tikzpicture}



\fill[black!5] (-8.5,4.5) -- (-8.5,0) -- (-4.15,1.65) -- (-3.5,5.5) -- cycle;
\draw[fill=black!5] (-6,0) circle (2.5);

\draw[blue,very thick] (-4.15,1.65) -- (-3.5,5.5);
\draw[blue,very thick] (-8.5,0) -- (-8.5,4.5);
\draw[black,very thick, dashed] (-8.5,0) -- (-4.15,1.65);

\node[anchor=west, xshift=-70pt, yshift=-5pt] at (-4.15,1.65) {\large line 3};
\node[anchor=west, xshift=-20pt, yshift=60pt] at (-4.15,1.65) {\large line 1};
\node[anchor=west, xshift=5pt, yshift=70pt] at (-8.5,0) {\large line 2};

\node[anchor=west, xshift=0pt, yshift=-83pt] at (-6,0) {\large (a)};

\fill[black!5] (-1.3,3.8) -- (-2.25,1.09) -- (1.4,2.0712) -- (0,2.8) -- (0.3,3.8) -- cycle;
\draw[fill=black!5] (0,0) circle (2.5);

\draw[blue,very thick] (1.4,2.0712) -- (0,2.8);
\draw[blue,very thick] (-2.25,1.09) -- (-1.3,3.8);
\draw[black,very thick, dashed] (1.4,2.0712) -- (-2.25,1.09);

\draw[black,very thick, dashed] (0,2.8) -- (0.3,3.8);
\draw[black,very thick, dashed] (0,2.8) -- (-1.5,3.3);

\node[anchor=west, xshift=40pt, yshift=40pt] at (-2.45,0.5) {\large line 3};
\node[anchor=west, xshift=-15pt, yshift=15pt] at (1,2.3) {\large line 1};
\node[anchor=west, xshift=-18pt, yshift=50pt] at (-2.45,0.5) {\large line 2};
\node[anchor=west, xshift=10pt, yshift=-12pt] at (-0.2,4.5) {\large line 4};
\node[anchor=west, xshift=-15pt, yshift=8pt] at (-0.5,3.2) {\large line 5};

\node[anchor=west, xshift=0pt, yshift=-83pt] at (0,0) {\large (b)};

\end{tikzpicture}}};
  \end{tikzpicture}%
  }
  \caption{Keyhole diagram with additional virtual lines.\figlabel{keyhole-lines}}
  \vspace*{-1.5em}
\end{figure}
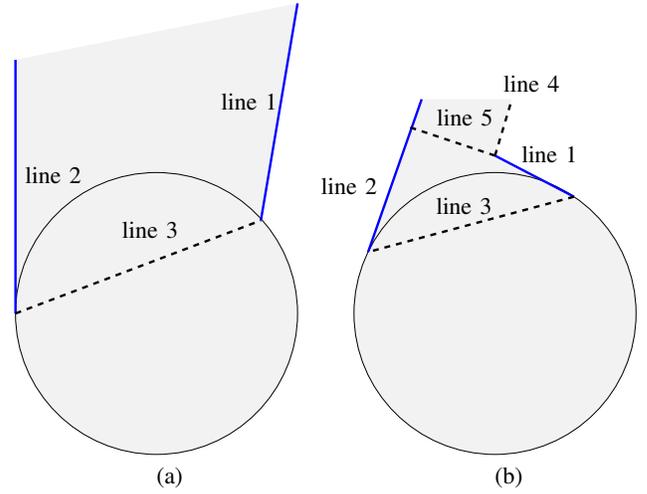

The synthesis process for the ZBF (i.e., training of the neural network) follows the technique presented in \cite{AA22PV}, which is a linear program (LP). The LP needs sampled sets, $\sampSet^u$ and $\sampSet^s$,
from the unsafe and safe regions, respectively. The unsafe points are sampled along the gap lines and circle edge, excluding the arc between the gap lines. The safe points are generated from the unsafe point by pushing them along the gradient inwards an $\epsilon$ distance. $\epsilon$ then should be set to a small value, e.g., 3$\%$ of the circle radius. 
The linear program for learning $\alpha_i$ and $b$ coefficients is 
\begin{equation}  \eqlabel{lp}
\begin{aligned}
    \min_{u} \quad & \vec{1}^T\alpha\\
    \text{s.t.} \quad & h(x_i)\leq -1,\, \forall i\in\set{l:x_l\in\sampSet^u}\\
                \quad & h(x_j)\geq +1,\, \forall j\in\set{l:x_l\in\sampSet^s}\\
                \quad & b\leq0,\ \alpha_k\geq 0,\, \forall k=1,\cdots,15
\end{aligned}
\end{equation}
where $\vec{1}=[1,\cdots,1]^T$ and $\alpha=[\alpha_1,\cdots,\alpha_{15}]$.
The coefficients $\alpha_i$ have a positivity constraint while the $b$ has a negativity constraint. Those constraints are needed so that the cost function acts as $L_1$ regulation, which promotes sparsity in the solution. Also, the value constant $\pm 1$ affects the scaling of the ZBF, much like for support vector machines. The synthesized {\keyhole} for the examples given in Fig.~\figref{keyhole-lines} are shown in Fig.~\figref{keyhole-levelset}.

The linear program was solved using Google OR-Tools \cite{ortools} in C++. For 2000 runs on the development  machine (Ubuntu 20.04, Intel i7-8750H CPU), the maximum, minimum, and average execution times were 1.6 ms, 0.71 ms, and 0.75 ms, respectively.

\subsubsection{{\keyhole} suitability as a CBF}
The {\keyhole} meets the requirements to be used as a CBF. It is monotonic across the boundary and differentiable everywhere in the positive region. The neural network in \eqref{nn} has no dead gradient, given that the terms are multiplicative combinations of the line and circle equations. Although due to the ReLU, the gradient may be non-smooth, it is not a problem for optimization, as subgradients can be used.

\begin{figure}[t]
    \vspace*{0.1in}
    \centering
    \begin{tikzpicture}[inner sep=0pt, outer sep=0pt]
        \node (fig_a) at (0in,0in)
        {\includegraphics[width=0.5\linewidth]{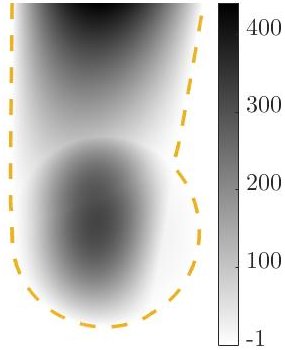}};
        \node[anchor=west, xshift=70pt] (fig_b) at (fig_a)
        {\includegraphics[width=0.5\linewidth]{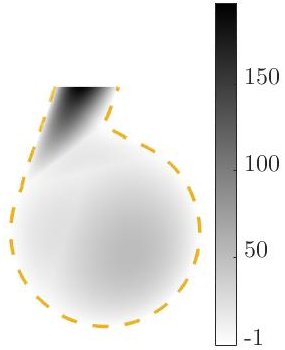}};

        \node[anchor=north, xshift=-15pt, yshift=-80pt] at (fig_a){(a)};
        \node[anchor=north, xshift=-15pt, yshift=-80pt] at (fig_b){(b)};
    \end{tikzpicture}
    
    \caption{{\keyhole} for configurations in Fig.~\figref{keyhole-lines}. The value of the ZBF is represented by the color map, and the zero level-set of the ZBF is depicted by the yellow dashed line. The unsafe region outside the ZBF boundary has a negative value.}
    \figlabel{keyhole-levelset}
    \vspace*{-1.5em}
\end{figure}

\subsection{Keyhole Control Barrier Function \label{sec:cbf}}

Since the domain of the {\keyhole} will be the position of the robot (excluding orientation), a single integrator model is assumed for the robot in CBF-QP. The reference control command is the instantaneous translational velocity of the robot, i.e., $u_r=[\nu_r\cos\theta,\nu_r\sin\theta]^T$. Again, the subscript $r$ denotes the outputs from the reference controller, which NMPC in this case. The calculated safe velocity commands by CBF-QP \eqref{cbf-qp}, $u=[\dot{x}_s,\dot{y}_s]^T$, are mapped to the robot commands using \eqref{w-qp} and \eqref{v-qp}. $\Delta\theta$ is the angle difference between the vectors $u_r$ and $u$ and is added to the current rotation rate to correct the angle difference. $k_{\omega}$ is a positive tunable parameter. The translational velocity is damped down proportional to the ratio of $\Delta\theta$ to a maximum angle $\theta_{max}$. If $\abs{\Delta\theta}\geq\theta_{max}$, the robot will only rotate. 
\begin{align}
    \omega &= \omega_r + k_{\omega}\Delta\theta \eqlabel{w-qp} \\ 
    \nu      &= \max\left(0,1-\frac{\abs{\Delta\theta}}{\theta_{max}}\right)\norm{u} \eqlabel{v-qp}
\end{align}

Overall, the {\saferGap} local planner is designed to maintain safety and passibility for nonholonomic mobile robots. From joined B\'{e}zier path planning, NMPC trajectory tracking with {\keyhole}, and control barrier function, safety guarantee is proved.

\section{Experiments \label{sec:exp}}
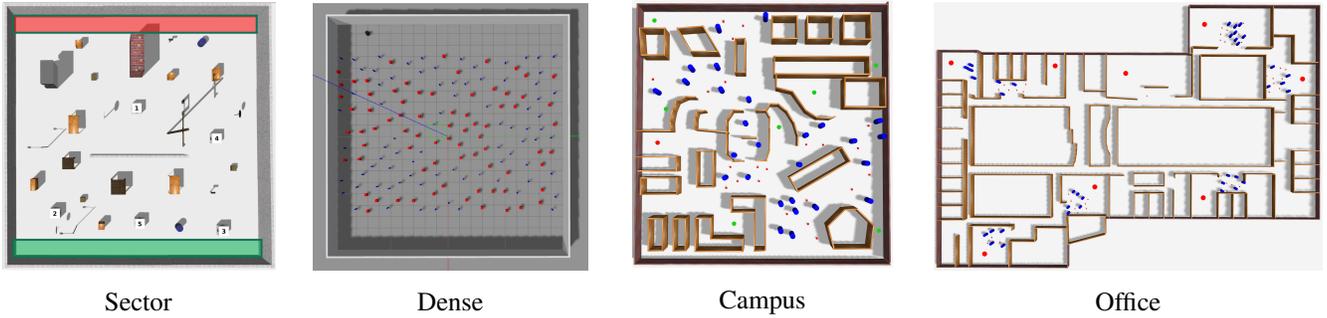
\begin{figure*}[htb]
\vspace*{0.06in}
  \centering
  \begin{minipage}{0.22\textwidth}
  \centering
  \begin{tikzpicture}[inner sep=0pt,outer sep=0pt]
    \node[anchor=south west] (comp1) at ($(0, 0)$)
    	{{\includegraphics[height=1.4in]{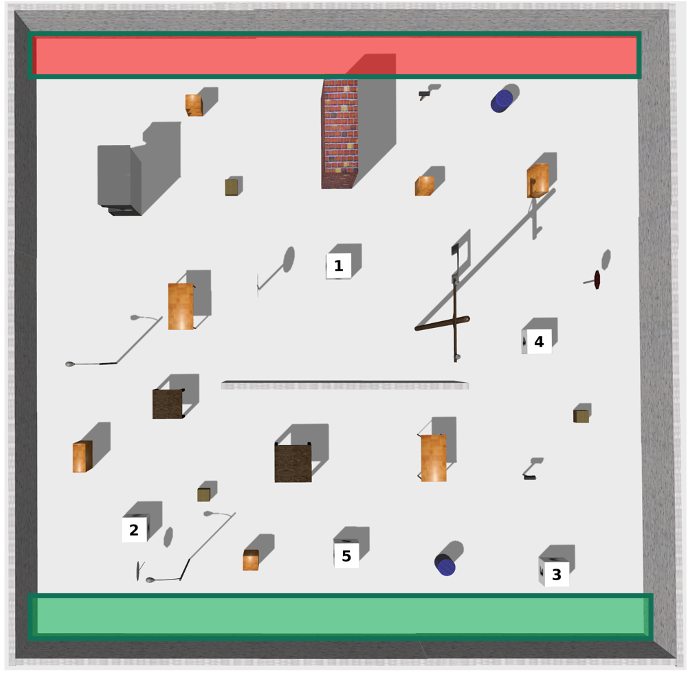}}};
    \node[anchor=north] at ($(comp1.south)+(0,-0.3)$) {\centering Sector};
  \end{tikzpicture}%
  \end{minipage}
  \hfill
  \begin{minipage}{0.22\textwidth}
  \centering
  \begin{tikzpicture}[inner sep=0pt,outer sep=0pt]
    \node[anchor=south west] (comp2) at ($(0, 0)$)
    	{{\includegraphics[height=1.4in]{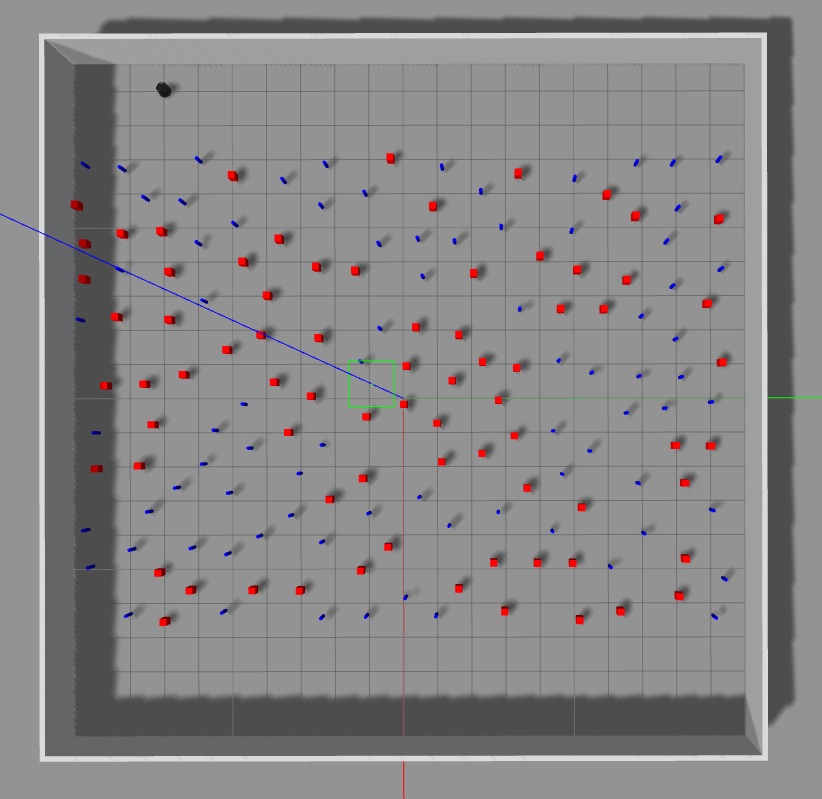}}};
    \node[anchor=north] at ($(comp2.south)+(0,-0.3)$) {\centering Dense};
  \end{tikzpicture}%
  \end{minipage}
  \hfill
  \begin{minipage}{0.22\textwidth}
  \centering
  \begin{tikzpicture}[inner sep=0pt,outer sep=0pt]
    \node[anchor=south west] (comp2) at ($(0, 0)$)
    	{{\includegraphics[height=1.4in]{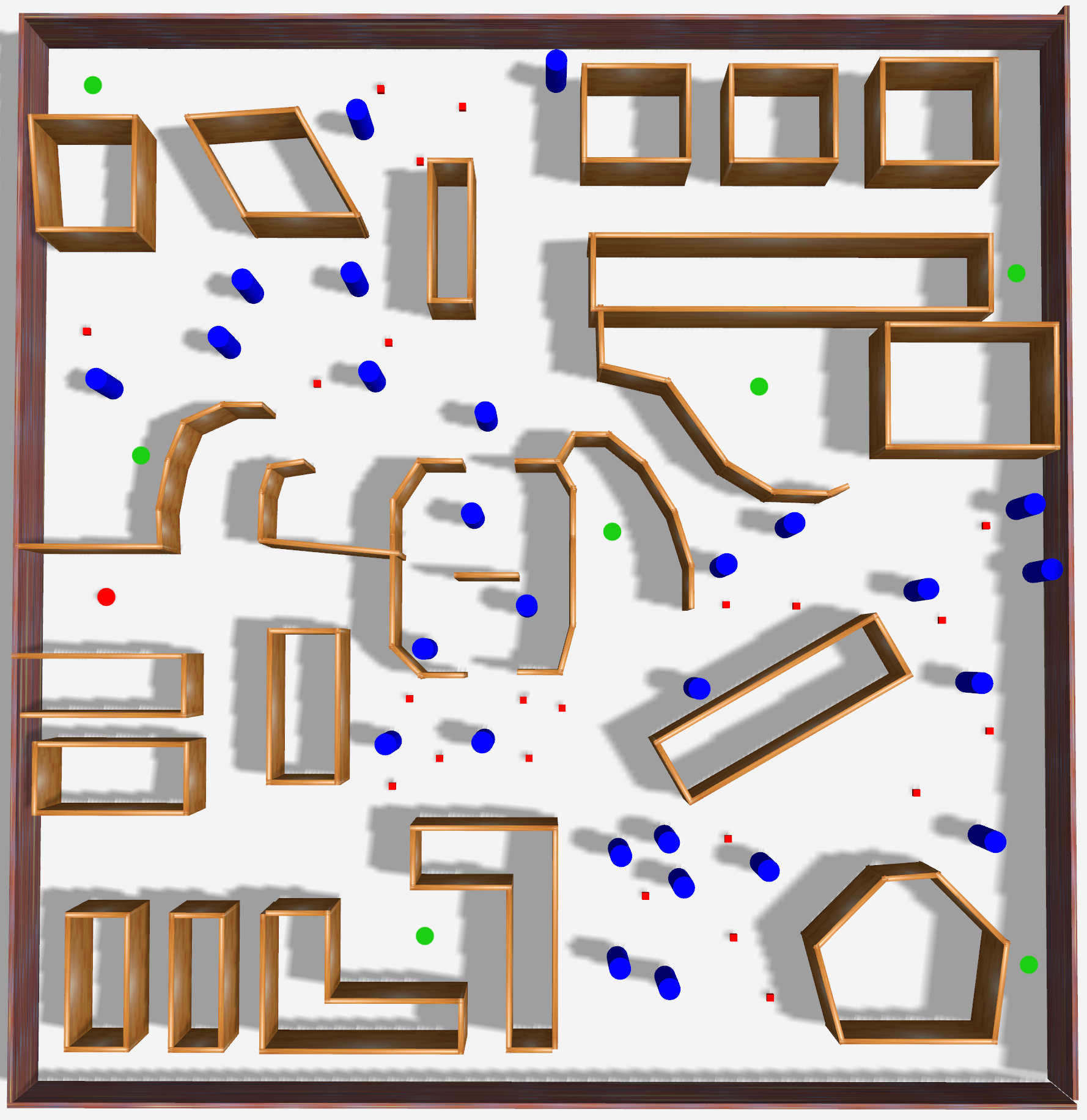}}};
    \node[anchor=north] at ($(comp2.south)+(0,-0.3)$) {\centering Campus};
  \end{tikzpicture}%
  \end{minipage}
  \hfill
  \begin{minipage}{0.3\textwidth}
  \centering
  \begin{tikzpicture}[inner sep=0pt,outer sep=0pt]
    \node[anchor=south west] (comp2) at ($(0, 0)$)
    	{{\includegraphics[height=1.4in]{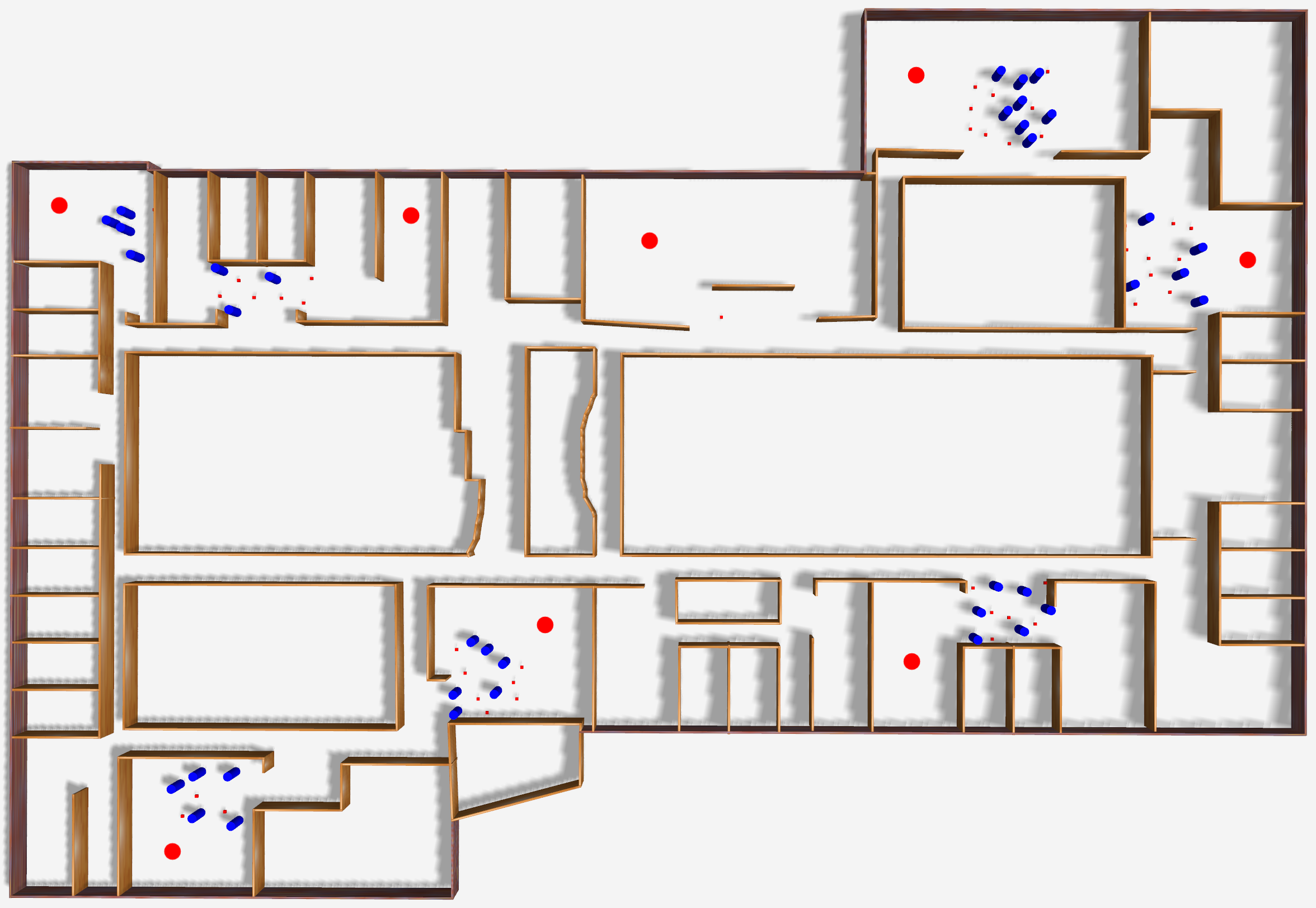}}};
    \node[anchor=north] at ($(comp2.south)+(0,-0.3)$) {\centering Office};
  \end{tikzpicture}%
  \end{minipage}
  \caption{Four simulation scenarios. In sector and campus worlds, start region/poses (red) and end region/poses (green) are labeled. In the dense world, robot navigates from top to bottom. In the office world, the start and goal poses are randomly chosen from the red points. \label{fig:sim}}
  \vspace*{-0.5em}
\end{figure*}

\begin{figure*}[!htb]
\begin{minipage}{0.3\textwidth}
\centering
\captionsetup{justification=centering}
\captionof{table}{ STDR Simulation Benchmark \\ ($1^{\text{st}}$ order nonholonomic model)\label{tab:sim_ben_s}}
\begin{tikzpicture}[inner sep=0pt,outer sep=0pt,scale=1, every node/.style={scale=1}]
    \node[anchor=west] (sim_stdr) at ($(0, 0pt)$)
    {
    \setlength{\tabcolsep}{4pt}
    \begin{tabular}{|c||ccc|}
    \hline 
    \textbf{Total} & Success & Abort & Collision \\ 
    \hline 
    PG & 100\%  & 0\%  & 0\%  \\ 
    SG & 100\%  & 0\%  & 0\%  \\ 
    \hline 
    \end{tabular}
    };

\end{tikzpicture}
\end{minipage}
\hfill
\begin{minipage}{0.3\textwidth}
\centering
\captionsetup{justification=centering}
\captionof{table}{ Gazebo Simulation Benchmark \\ ($2^{\text{nd}}$ order nonholonomic model)\label{tab:sim_ben_g}}
\begin{tikzpicture}[inner sep=0pt,outer sep=0pt,scale=1, every node/.style={scale=1}]
    \node[anchor=west] (sim_stdr) at ($(0, 0pt)$)
    {
    \setlength{\tabcolsep}{4pt}
    \begin{tabular}{|c||ccc|}
    \hline 
    \textbf{Total} & Success & Abort & Collision \\ 
    \hline 
    PG & 91\%  & 8\%  & 1\%  \\ 
    SG & 100\%  & 0\%  & 0\%  \\
    \hline 
    \end{tabular}
    };

\end{tikzpicture}
\end{minipage}
\hfill
\begin{minipage}{0.38\textwidth}
\centering
  \scalebox{0.75}{
  \begin{tikzpicture}[inner sep=0pt,outer sep=0pt]
    \node[anchor=south west] at ($(0, 0)$)
    	{{\includegraphics[height=1.1in]{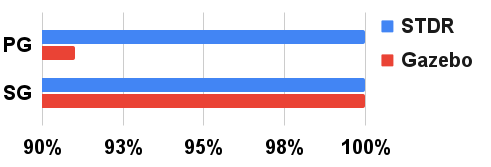}}};
  \end{tikzpicture}%
  }
  \caption{Success rates comparison between PG and SG. \label{fig:sim_comp}}
\end{minipage}
\vspace*{-1em}
\end{figure*}

\begin{figure*}[!htb]
\vspace*{-0.5em}
\begin{minipage}{\textwidth}
\centering
\vspace*{0.1in}
\captionof{table}{ Simulation results in 4 Gazebo scenarios \label{tab:sim_ben_r}}
\begin{tikzpicture}[inner sep=0pt,outer sep=0pt,scale=1, every node/.style={scale=1}]

	\node[anchor=north west] (sim_dense) at (0, 0pt)
    {
    \setlength{\tabcolsep}{4pt}
    \begin{tabular}{|c||ccc||ccc||ccc||ccc|}
    \hline 
    & \multicolumn{3}{|c||}{\textbf{Sector}} & \multicolumn{3}{|c||}{\textbf{Dense}} & \multicolumn{3}{|c||}{\textbf{Campus}} & \multicolumn{3}{|c|}{\textbf{Office}} \\
    \hline
     & Success & Abort & Collision & Success & Abort & Collision & Success & Abort & Collision & Success & Abort & Collision \\ 
    \hline 
    PG & 100\%  & 0\%  & 0\%  & 100\%  & 0\%  & 0\% & 84\%  & 16\%  & 0\% & 80\%  & 16\%  & 4\% \\ 
    SG & 100\%  & 0\%  & 0\% & 100\%  & 0\%  & 0\% & 100\%  & 0\%  & 0\% & 100\%  & 0\%  & 0\%  \\ 
    \hline 
    \end{tabular}
    };

\end{tikzpicture}
\end{minipage}
\end{figure*}

\begin{figure*}[!htb]
  \centering
  \begin{minipage}{0.32\textwidth}
  \centering
  \begin{tikzpicture}[inner sep=0pt,outer sep=0pt]
    \node[anchor=south west] (comp1) at ($(0, 0)$)
    	{{\includegraphics[height=1.25in]{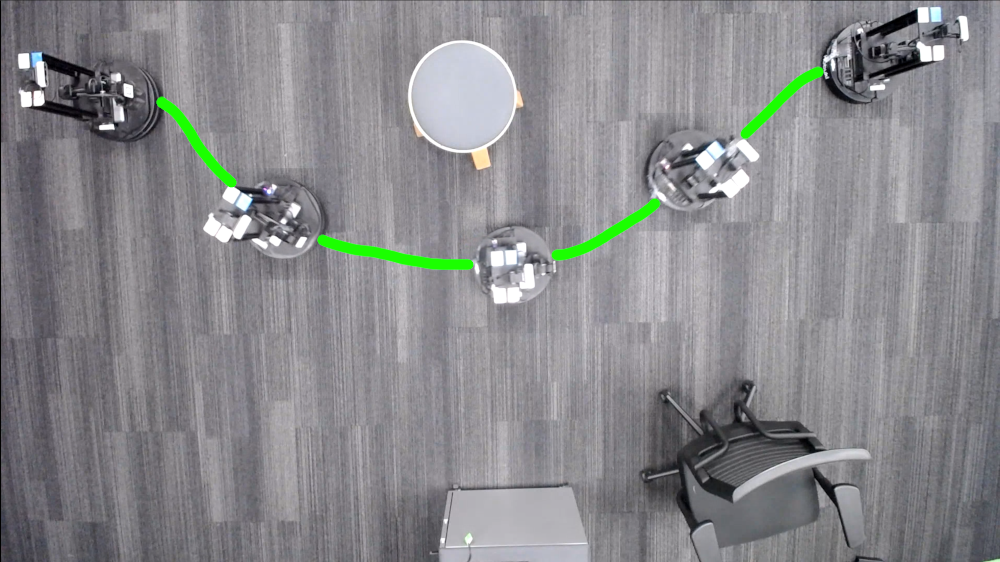}}};
    \node[anchor=north] at ($(comp1.south)+(0,-0.3)$) {\centering Low Density};

    \node[anchor=north,color=yellow] at ($(comp1.south)+(-2.3,1.8)$) {\centering \Large \textbf{Start}};
    \node[anchor=north,color=yellow] at ($(comp1.south)+(2.3,1.8)$) {\centering \Large \textbf{Goal}};
  \end{tikzpicture}%
  \end{minipage}
  \hfill
  \begin{minipage}{0.32\textwidth}
  \centering
  \begin{tikzpicture}[inner sep=0pt,outer sep=0pt]
    \node[anchor=south west] (comp2) at ($(0, 0)$)
    	{{\includegraphics[height=1.25in]{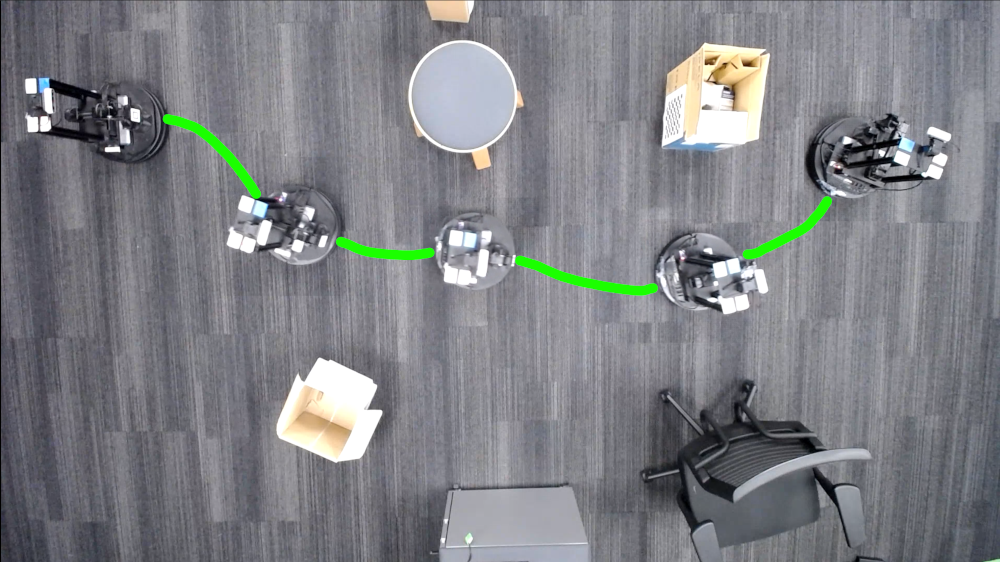}}};
    \node[anchor=north] at ($(comp2.south)+(0,-0.3)$) {\centering Medium Density};

    \node[anchor=north,color=yellow] at ($(comp1.south)+(-2.3,1.8)$) {\centering \Large \textbf{Start}};
    \node[anchor=north,color=yellow] at ($(comp1.south)+(2.3,1.8)$) {\centering \Large \textbf{Goal}};
  \end{tikzpicture}%
  \end{minipage}
  \hfill
  \begin{minipage}{0.32\textwidth}
  \centering
  \begin{tikzpicture}[inner sep=0pt,outer sep=0pt]
    \node[anchor=south west] (comp2) at ($(0, 0)$)
    	{{\includegraphics[height=1.25in]{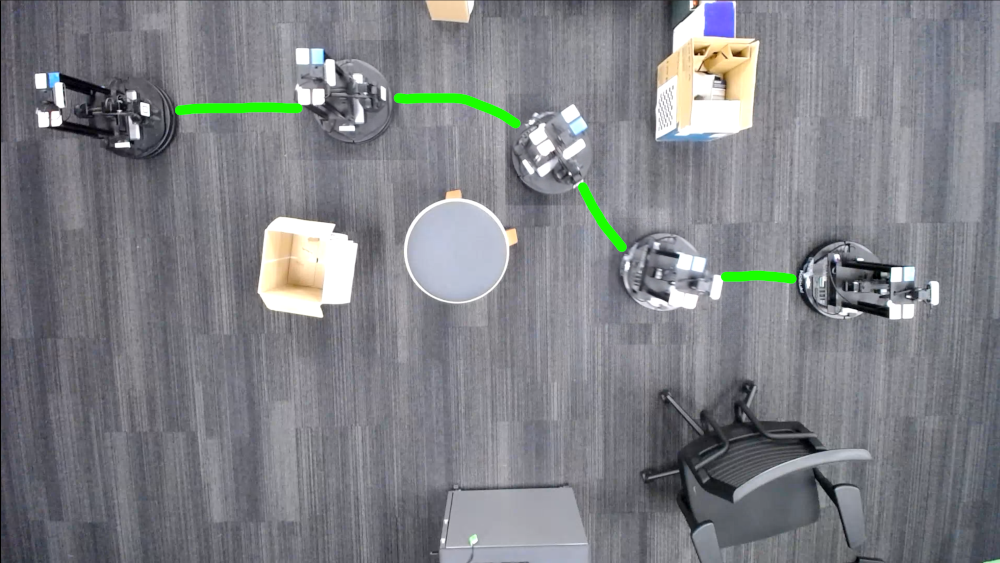}}};
    \node[anchor=north] at ($(comp2.south)+(0,-0.3)$) {\centering High Density};

    \node[anchor=north,color=yellow] at ($(comp1.south)+(-2.3,1.8)$) {\centering \Large \textbf{Start}};
    \node[anchor=north,color=yellow] at ($(comp1.south)+(2.3,2.5)$) {\centering \Large \textbf{Goal}};
  \end{tikzpicture}%
  \end{minipage}
  \caption{Real experiment topview. Three environment densities are shown. The green paths are the robot's real traces. Robots start from left to the right. \label{fig:real}}
  \vspace*{-0.5em}
\end{figure*}

To test navigation performance with {\saferGap}, simulation benchmark and real robot experiments are conducted.

\subsection{Simulation Benchmark}

\subsubsection{Simulation Configuration}
We benchmarked {\saferGap} local planer in ROS with move\_base hierarchical navigation system.  The benchmark is performed in both STDR and Gazebo simulators. STDR uses $1^{\text{st}}$ order circular nonholonomic model with 360$^\circ$ Field-of-View (FoV) laser scanner. In Gazebo, Turtlebot is used as the $2^{\text{nd}}$ order nonholonomic mobile robot with limited 60$^\circ$ FoV. We setup four different scenarios \cite{Smith2020} in Fig.~\ref{fig:sim} for benchmark: sector, dense, campus, and office. They simulate multiple navigation environments, e.g. hallway, open area with obstacles, campus roads and etc. The obstacles are randomly spawned in the scenarios with a minimum distance to each other as 1m. Robot start and end poses are also randomly chosen in the designated areas. Ground truth robot locations are used for navigation.  We have 50 Monte Carlo runs for each scenario to quantitatively compare {\saferGap} (SG) navigation performance with {\pGap} (PG) which serves as the baseline method. Radial extension and projection operator are enabled in PG for nonholonomic robot.

\subsubsection{Evaluation Metric}

Success, abort and collision rates are collected for each planner. Success means that the robot can successfully reach the goal. Abort represents that the robot cannot have any possible plan to the goal after recovery behavior. Collision is counted whenever a colliding happens.  

\subsubsection{Simulation Results}

The overall simulation results of STDR and Gazebo are in Table~\ref{tab:sim_ben_s} and \ref{tab:sim_ben_g}. The comparison of success rates is in Fig.~\ref{fig:sim_comp}. Both PG and SG have 100\% success rates in STDR with full sensing of the environments. However, when simulating $2^{\text{nd}}$ order nonholonomic model with limited FoV, PG's success rate drops 9\% including 8\% aborts and 1\% collisions. 

From the full results of 4 Gazebo scenarios in Table~\ref{tab:sim_ben_r}, PG does not perform well in the hallway and campus roads. The robot navigates back and forth, and is stuck to find successful paths to the goal. The projection operator in PG conservatively keep safety by sacrificing passibility. Due to the limited FoV, the PG still has collisions when all nonholonomic extensions are enabled. Our proposed local planner SG not only maintains safety, but also leverages the line-of-sight visible gaps for passibility. This comparison demonstrates safety guarantee for nonholonomic mobile robots in the design of {\saferGap}.

\subsubsection{Computational Efficiency}

The benchmark is performed on a workstation with Intel i7-8700. CasADi optimization framework is used for solving NMPC. We set the number of horizon $N=6$. The total time of each control loop is averaged as $\sim 75ms$, including B\'{e}zier path synthesis, keyhole generation, NMPC optimization, and CBF-QP. Real time application is achievable.

\subsection{Real Experiments}

The simulation benchmark quantitatively compares {\saferGap} navigation performance with {\pGap}, and presents the outperforming results of our proposed work. In this section, the planner is applied on the real platform, LoCoBot, to navigate through unknown environments. It has a Kobuki base with nonholonomic dynamics. Robot's odometry provides the pose information. RealSense D435i depth camera initially produces depth images that are converted to laserscan measurements through a ROS package depthimage\_to\_laserscan \cite{depthToLaser}.

We test {\saferGap} local planner in five scenarios with different obstacle densities from low to high in Fig.~\ref{fig:real}. The robot navigation traces are depicted on the top-view figures. For each configuration, 2 runs are repeated to show consistent results. We have 100\% success rate in totally 10 trials, which is the same as simulation. Therefore, {\saferGap} local planner is applicable on the real nonholonomic robots to maintain navigation safety and passibility.

\section{Conclusion \label{sec:conclusion}}
The proposed {\saferGap} local planner designs a safe navigation policy for nonholonomic mobile robots. It generates smooth joined B\'{e}zier paths in the collision free space defined by gaps to guarantee safety. We performs NMPC to track the reference path considering nonholonomic dynamics. A synthesized {\keyhole} for the free space is integrated as a safety constraint to prevent collision with obstacles. At the end, keyhole control barrier function provides additional safety at the lowest level of navigation hierarchy. From the simulation benchmark and real experiments, {\saferGap} is demonstrated to achieve safe navigation for nonholonomic robots without losing gap passibility. In the future work, different robot dynamics and environment complexities should be tested to evaluate the robustness of {\saferGap}.

\balance

\bibliographystyle{IEEEtran}
\bibliography{ref}

\begin{thebibliography}{10}
\providecommand{\url}[1]{#1}
\csname url@rmstyle\endcsname
\providecommand{\newblock}{\relax}
\providecommand{\bibinfo}[2]{#2}
\providecommand\BIBentrySTDinterwordspacing{\spaceskip=0pt\relax}
\providecommand\BIBentryALTinterwordstretchfactor{4}
\providecommand\BIBentryALTinterwordspacing{\spaceskip=\fontdimen2\font plus
\BIBentryALTinterwordstretchfactor\fontdimen3\font minus
  \fontdimen4\font\relax}
\providecommand\BIBforeignlanguage[2]{{%
\expandafter\ifx\csname l@#1\endcsname\relax
\typeout{** WARNING: IEEEtran.bst: No hyphenation pattern has been}%
\typeout{** loaded for the language `#1'. Using the pattern for}%
\typeout{** the default language instead.}%
\else
\language=\csname l@#1\endcsname
\fi
#2}}

\bibitem{pgap}
R.~Xu, S.~Feng, and P.~A. Vela, ``Potential gap: A gap-informed reactive policy
  for safe hierarchical navigation,'' \emph{IEEE Robotics and Automation
  Letters}, vol.~6, no.~4, pp. 8325--8332, 2021.

\bibitem{bezierbook}
\BIBentryALTinterwordspacing
H.~Prautzsch, W.~Boehm, and M.~Paluszny, \emph{B{\'{e}}zier and B-Spline
  Techniques}.\hskip 1em plus 0.5em minus 0.4em\relax Springer Berlin
  Heidelberg, 2002. [Online]. Available:
  \url{https://doi.org/10.1007/978-3-662-04919-8}
\BIBentrySTDinterwordspacing

\bibitem{Smith2020}
J.~S. Smith, S.~Feng, F.~Lyu, and P.~A. Vela, \emph{Real-Time Egocentric
  Navigation Using 3D Sensing}.\hskip 1em plus 0.5em minus 0.4em\relax Cham:
  Springer International Publishing, 2020, pp. 431--484.

\bibitem{followthegap}
V.~Sezer and M.~Gokasan, ``A novel obstacle avoidance algorithm:follow the gap
  method,'' in \emph{RAS}, vol.~60, no.~9, July 2012, pp. 1123--1134.

\bibitem{CG}
M.~{Mujahad}, D.~{Fischer}, B.~{Mertsching}, and H.~{Jaddu}, ``Closest gap
  based (cg) reactive obstacle avoidance navigation for highly cluttered
  environments,'' in \emph{IROS}, 2010, pp. 1805--1812.

\bibitem{TCG}
M.~{Mujahed}, H.~{Jaddu}, D.~{Fischer}, and B.~{Mertsching}, ``Tangential
  closest gap based (tcg) reactive obstacle avoidance navigation for cluttered
  environments,'' in \emph{SSRR}, 2013, pp. 1--6.

\bibitem{SG}
M.~{Mujahed}, D.~{Fischer}, and B.~{Mertsching}, ``Safe gap based (sg) reactive
  navigation for mobile robots,'' in \emph{ECMR}, 2013, pp. 325--330.

\bibitem{AG}
M.~{Mujahed} and B.~{Mertsching}, ``The admissible gap (ag) method for reactive
  collision avoidance,'' in \emph{ICRA}, 2017, pp. 1916--1921.

\bibitem{SmEtAl[2020]egoTEB}
J.~S. {Smith}, R.~{Xu}, and P.~{Vela}, ``{egoTEB: Egocentric, Perception Space
  Navigation Using Timed-Elastic-Bands},'' in \emph{ICRA}, 2020, pp.
  2703--2709.

\bibitem{dijkstra}
\BIBentryALTinterwordspacing
E.~W. Dijkstra, ``A note on two problems in connexion with graphs,''
  \emph{Numer. Math.}, vol.~1, no.~1, p. 269–271, dec 1959. [Online].
  Available: \url{https://doi.org/10.1007/BF01386390}
\BIBentrySTDinterwordspacing

\bibitem{Astar}
P.~E. Hart, N.~J. Nilsson, and B.~Raphael, ``A formal basis for the heuristic
  determination of minimum cost paths,'' \emph{IEEE Transactions on Systems
  Science and Cybernetics}, vol.~4, no.~2, pp. 100--107, 1968.

\bibitem{Dstar}
A.~Stentz, ``Optimal and efficient path planning for partially-known
  environments,'' in \emph{Proceedings of the 1994 IEEE International
  Conference on Robotics and Automation}, 1994, pp. 3310--3317 vol.4.

\bibitem{DstarLite}
S.~Koenig and M.~Likhachev, ``Fast replanning for navigation in unknown
  terrain,'' \emph{IEEE Transactions on Robotics}, vol.~21, no.~3, pp.
  354--363, 2005.

\bibitem{LIKHACHEV20081613}
\BIBentryALTinterwordspacing
M.~Likhachev, D.~Ferguson, G.~Gordon, A.~Stentz, and S.~Thrun, ``Anytime search
  in dynamic graphs,'' \emph{Artificial Intelligence}, vol. 172, no.~14, pp.
  1613--1643, 2008. [Online]. Available:
  \url{https://www.sciencedirect.com/science/article/pii/S000437020800060X}
\BIBentrySTDinterwordspacing

\bibitem{prm}
L.~Kavraki, M.~Kolountzakis, and J.-C. Latombe, ``Analysis of probabilistic
  roadmaps for path planning,'' \emph{IEEE Transactions on Robotics and
  Automation}, vol.~14, no.~1, pp. 166--171, 1998.

\bibitem{lazyPRM}
R.~Bohlin and L.~Kavraki, ``Path planning using lazy prm,'' in
  \emph{Proceedings 2000 ICRA. Millennium Conference. IEEE International
  Conference on Robotics and Automation. Symposia Proceedings (Cat.
  No.00CH37065)}, vol.~1, 2000, pp. 521--528 vol.1.

\bibitem{rrt}
E.~Frazzoli, M.~Dahleh, and E.~Feron, ``Real-time motion planning for agile
  autonomous vehicles,'' in \emph{Proceedings of the 2001 American Control
  Conference. (Cat. No.01CH37148)}, vol.~1, 2001, pp. 43--49 vol.1.

\bibitem{Otte2014RRTXRM}
M.~W. Otte and E.~Frazzoli, ``Rrtx: Real-time motion planning/replanning for
  environments with unpredictable obstacles,'' in \emph{Workshop on the
  Algorithmic Foundations of Robotics}, 2014.

\bibitem{eb}
S.~Quinlan and O.~Khatib, ``Elastic bands: connecting path planning and
  control,'' in \emph{[1993] Proceedings IEEE International Conference on
  Robotics and Automation}, 1993, pp. 802--807 vol.2.

\bibitem{DWA}
D.~{Fox}, W.~{Burgard}, and S.~{Thrun}, ``The dynamic window approach to
  collision avoidance,'' \emph{RA-M}, vol.~4, no.~1, pp. 23--33, March 1997.

\bibitem{TEB}
C.~{Rösmann}, F.~{Hoffmann}, and T.~{Bertram}, ``Timed-elastic-bands for
  time-optimal point-to-point nonlinear model predictive control,'' in
  \emph{ECC}, July 2015, pp. 3352--3357.

\bibitem{APF1985}
O.~{Khatib}, ``Real-time obstacle avoidance for manipulators and mobile
  robots,'' in \emph{ICRA}, 1985, pp. 500--505.

\bibitem{APF_numerical}
J.~{Barraquand}, B.~{Langlois}, and J.~. {Latombe}, ``Numerical potential field
  techniques for robot path planning,'' \emph{T-SMC}, vol.~22, no.~2, pp.
  224--241, 1992.

\bibitem{APF_newfunction}
S.~S. {Ge} and Y.~J. {Cui}, ``New potential functions for mobile robot path
  planning,'' \emph{IEEE Trans. on Rob. and Aut.}, vol.~16, no.~5, pp.
  615--620, 2000.

\bibitem{APFSimAnnel}
M.~{Park}, J.~{Jeon}, and M.~{Lee}, ``Obstacle avoidance for mobile robots
  using artificial potential field approach with simulated annealing,'' in
  \emph{ISIE}, vol.~3, 2001, pp. 1530--1535 vol.3.

\bibitem{APFRegression}
G.~{Li}, A.~{Yamashita}, H.~{Asama}, and Y.~{Tamura}, ``An efficient improved
  artificial potential field based regression search method for robot path
  planning,'' in \emph{ICMA}, 2012, pp. 1227--1232.

\bibitem{Choi2008PathPB}
J.~wung Choi, R.~E. Curry, and G.~H. Elkaim, ``Path planning based on
  b{\'e}zier curve for autonomous ground vehicles,'' \emph{Advances in
  Electrical and Electronics Engineering - IAENG Special Edition of the World
  Congress on Engineering and Computer Science 2008}, pp. 158--166, 2008.

\bibitem{ELHOSENY2018339}
\BIBentryALTinterwordspacing
M.~Elhoseny, A.~Tharwat, and A.~E. Hassanien, ``Bezier curve based path
  planning in a dynamic field using modified genetic algorithm,'' \emph{Journal
  of Computational Science}, vol.~25, pp. 339--350, 2018. [Online]. Available:
  \url{https://www.sciencedirect.com/science/article/pii/S1877750317308906}
\BIBentrySTDinterwordspacing

\bibitem{s21072460}
\BIBentryALTinterwordspacing
H.~A. Satai, M.~M.~A. Zahra, Z.~I. Rasool, R.~S. Abd-Ali, and C.~I. Pruncu,
  ``Bézier curves-based optimal trajectory design for multirotor uavs with
  any-angle pathfinding algorithms,'' \emph{Sensors}, vol.~21, no.~7, 2021.
  [Online]. Available: \url{https://www.mdpi.com/1424-8220/21/7/2460}
\BIBentrySTDinterwordspacing

\bibitem{Zheng2020BezierTraj}
\BIBentryALTinterwordspacing
L.~Zheng, P.~Zeng, W.~Yang, Y.~Li, and Z.~Zhan, ``Bézier curve-based
  trajectory planning for autonomous vehicles with collision avoidance,''
  \emph{IET Intelligent Transport Systems}, vol.~14, no.~13, pp. 1882--1891,
  2020. [Online]. Available:
  \url{https://ietresearch.onlinelibrary.wiley.com/doi/abs/10.1049/iet-its.2020.0355}
\BIBentrySTDinterwordspacing

\bibitem{HASSANI2018305}
\BIBentryALTinterwordspacing
V.~Hassani and S.~V. Lande, ``Path planning for marine vehicles using
  b\'{e}zier curves,'' \emph{IFAC-PapersOnLine}, vol.~51, no.~29, pp. 305--310,
  2018, 11th IFAC Conference on Control Applications in Marine Systems,
  Robotics, and Vehicles CAMS 2018. [Online]. Available:
  \url{https://www.sciencedirect.com/science/article/pii/S2405896318322092}
\BIBentrySTDinterwordspacing

\bibitem{BryceUAVBezier}
B.~Ingersoll, J.~Ingersoll, P.~DeFranco, and A.~Ning, ``Uav path-planning using
  bezier curves and a receding horizon approach,'' in \emph{AIAA Modeling and
  Simulation Technologies Conference}, 06 2016.

\bibitem{bgap}
S.~Feng, Z.~Zhou, J.~S. Smith, M.~Asselmeier, Y.~Zhao, and P.~A. Vela,
  ``{GPF-BG}: A hierarchical vision-based planning framework for safe
  quadrupedal navigation,'' to appear in 2023 IEEE International Conference on
  Robotics and Automation (ICRA), 2023.

\bibitem{mpc-book}
F.~Borrelli, A.~Bemporad, and M.~Morari, \emph{Predictive Control for Linear
  and Hybrid Systems}.\hskip 1em plus 0.5em minus 0.4em\relax Cambridge
  University Press, 2017.

\bibitem{mpc-wheeled}
\BIBentryALTinterwordspacing
T.~P. Nascimento, C.~E.~T. D{\'{o}}rea, and L.~M.~G. Gon{\c{c}}alves,
  ``Nonlinear model predictive control for trajectory tracking of nonholonomic
  mobile robots,'' \emph{International Journal of Advanced Robotic Systems},
  vol.~15, no.~1, p. 172988141876046, Jan. 2018. [Online]. Available:
  \url{https://doi.org/10.1177/1729881418760461}
\BIBentrySTDinterwordspacing

\bibitem{mpcTEB}
C.~R\"smann, A.~Makarow, and T.~Bertram, ``Online motion planning based on
  nonlinear model predictive control with non-euclidean rotation groups,'' in
  \emph{2021 European Control Conference (ECC)}, 2021, pp. 1583--1590.

\bibitem{AA17TP}
A.~D. Ames, X.~Xu, J.~W. Grizzle, and P.~Tabuada, ``Control barrier function
  based quadratic programs for safety critical systems,'' \emph{IEEE
  Transactions on Automatic Control}, vol.~62, no.~8, pp. 3861--3876, 2017.

\bibitem{HM22ET}
H.~Almubarak, N.~Sadegh, and E.~A. Theodorou, ``Safety embedded control of
  nonlinear systems via barrier states,'' \emph{IEEE Control Systems Letters},
  vol.~6, pp. 1328--1333, 2022.

\bibitem{JZ21KS}
J.~Zeng, B.~Zhang, and K.~Sreenath, ``Safety-critical model predictive control
  with discrete-time control barrier function,'' in \emph{2021 American Control
  Conference (ACC)}, 2021, pp. 3882--3889.

\bibitem{AAsos}
Y.~Chen, M.~Jankovic, M.~Santillo, and A.~D. Ames, ``Backup control barrier
  functions: Formulation and comparative study,'' in \emph{2021 60th IEEE
  Conference on Decision and Control (CDC)}, 2021, pp. 6835--6841.

\bibitem{SB17CT}
S.~Bansal, M.~Chen, S.~Herbert, and C.~J. Tomlin, ``Hamilton-jacobi
  reachability: A brief overview and recent advances,'' in \emph{2017 IEEE 56th
  Annual Conference on Decision and Control (CDC)}, 2017, pp. 2242--2253.

\bibitem{MC18CT}
\BIBentryALTinterwordspacing
M.~Chen and C.~J. Tomlin, ``Hamilton{\textendash}jacobi reachability: Some
  recent theoretical advances and applications in unmanned airspace
  management,'' \emph{Annual Review of Control, Robotics, and Autonomous
  Systems}, vol.~1, no.~1, pp. 333--358, May 2018. [Online]. Available:
  \url{https://doi.org/10.1146/annurev-control-060117-104941}
\BIBentrySTDinterwordspacing

\bibitem{Kousik2020}
\BIBentryALTinterwordspacing
S.~Kousik, S.~Vaskov, F.~Bu, M.~Johnson-Roberson, and R.~Vasudevan, ``Bridging
  the gap between safety and real-time performance in receding-horizon
  trajectory design for mobile robots,'' \emph{The International Journal of
  Robotics Research}, vol.~39, no.~12, pp. 1419--1469, Aug. 2020. [Online].
  Available: \url{https://doi.org/10.1177/0278364920943266}
\BIBentrySTDinterwordspacing

\bibitem{FasTrack}
M.~Chen, S.~L. Herbert, H.~Hu, Y.~Pu, J.~F. Fisac, S.~Bansal, S.~Han, and C.~J.
  Tomlin, ``Fastrack:a modular framework for real-time motion planning and
  guaranteed safe tracking,'' \emph{IEEE Transactions on Automatic Control},
  vol.~66, no.~12, pp. 5861--5876, 2021.

\bibitem{1025398}
R.~Olfati-Saber, ``Near-identity diffeomorphisms and exponential /spl
  epsi/-tracking and /spl epsi/-stabilization of first-order nonholonomic se(2)
  vehicles,'' in \emph{Proceedings of the 2002 American Control Conference
  (IEEE Cat. No.CH37301)}, vol.~6, 2002, pp. 4690--4695 vol.6.

\bibitem{AA22PV}
\BIBentryALTinterwordspacing
A.~Abuaish, M.~Srinivasan, and P.~A. Vela, ``Geometry of radial basis neural
  networks for safety biased approximation of unsafe regions,'' 2022. [Online].
  Available: \url{https://arxiv.org/abs/2210.05596}
\BIBentrySTDinterwordspacing

\bibitem{ortools}
\BIBentryALTinterwordspacing
L.~Perron and V.~Furnon, ``Or-tools,'' Google. [Online]. Available:
  \url{https://developers.google.com/optimization/}
\BIBentrySTDinterwordspacing

\bibitem{depthToLaser}
\BIBentryALTinterwordspacing
(2020) depthimage\_to\_laserscan - ros wiki. [Online]. Available:
  \url{http://wiki.ros.org/depthimage\_to\_laserscan}
\BIBentrySTDinterwordspacing

\end{thebibliography}

\end{document}